\documentclass[journal]{IEEEtran}
\usepackage{blindtext}
\usepackage{graphicx}
\usepackage{multirow}
\usepackage{booktabs}
\usepackage{xcolor}
\usepackage{amssymb}
\usepackage{mathtools}
\usepackage{hyperref}
\usepackage{url}

\begin{document}

\title{FragNet: Writer Identification using \\ Deep Fragment Networks}

\author{Sheng~He,
        Lambert~Schomaker,~\IEEEmembership{Senior,~IEEE,}
       
\thanks{
S.He is with the Boston Children's Hospital and Harvard Medical School, Harvard University, 300 Longwood Ave., Boston, MA, USA (e-mail: heshengxgd@gmail.com)}% <-this % stops a space
\thanks{L.Schomaker is with the Bernoulli Institute for Mathematics, Computer Science and Artificial Intelligence, University of Groningen, Groningen, PO Box 407, 9700 AK, The Netherlands 
(e-mail: l.r.b.schomaker@rug.nl).}% <-this % stops a space
}

\maketitle

\begin{abstract}
Writer identification based on a small amount of text is a challenging problem.
In this paper, we propose a new benchmark study for writer identification based on word or text block images which approximately contain one word.
In order to extract powerful features on these word images, a deep neural network, named FragNet, is proposed.
The FragNet has two pathways: feature pyramid which is used to extract feature maps and fragment pathway which is trained to predict the writer identity based on fragments extracted from the input image and the feature maps on the feature pyramid.
We conduct experiments on four benchmark datasets, which show that our proposed method can generate efficient and robust deep representations for writer identification based on both word and page images.

\end{abstract}


\IEEEpeerreviewmaketitle
\section{Introduction}

Writer identification is a typical pattern recognition problem in forensic science and has been studied for many years.
It has potential applications in historical~\cite{brink2012writer} and forensic~\cite{pervouchine2007extraction} document analysis.
Forensic document examiners usually use certain features of handwritten texts to determine the authorship.
Therefore, traditional methods focus on extracting handcrafted features to capture the writing style information of a given handwritten document~\cite{bulacu2007text,siddiqi2010text,newell2014writer}.
Handcrafted features usually have been designed to describe specific attributes of the natural handwriting style of a writer, such as slant~\cite{bulacu2007text} and curvature~\cite{siddiqi2010text}.
Due to the fact that handcrafted features capture the statistical information of handwriting style which need more information to obtain a good performance, a system using handcrafted features for writer identification usually requires a large amount of handwritten text, such as a paragraph or a text block with several sentences~\cite{bulacu2007text,khan2018dissimilarity}.

The traditional writer identification methods focus on extracting handcrafted or deep learned features on the whole document images.
Each writer usually has two samples: one is used for training and the other is used for evaluation. 
However, as discussed in our previous work~\cite{he2019deep}, in the digital era,  a real-world writer identification system should recognize the author based on a very small amount of handwritten text since people are preferring typing on keyboards instead of writing on papers.
Thus, we propose a new protocol for writer identification based on the traditional benchmark datasets:
identifying the writer based on word or text-block images segmented on the whole handwritten documents.
Following the traditional writer identification protocol, word or text-block images segmented on one page of each writer are used for modeling the handwriting style and text-block images segmented on other documents from the same writer are used for testing.
Therefore, it can be used for both word-based and page-based writer identification tasks. 
Word-based writer identification is the problem of identifying the writer based on word images while page-based writer identification relies on global features computed by aggregating local features extracted on word or text-block images. 

Writer identification based on word images is more challenging than writer identification based on page images since the writer-related style information in word images is limited for modeling the writer's handwriting style. 
Our previous work~\cite{he2019deep} studies on writer identification based on word images using Convolutional neural networks (CNNs) with two branches for different tasks, which requires extra annotations for the auxiliary task. 
Therefore, it can be only applied on these word images which carry word labels.
In this paper, we propose a new method, which recognizes the writer identity based on word images using deep learning~\cite{lecun1998gradient,krizhevsky2012imagenet} without any extra labels.
 CNNs can learn deep abstract and high level features on a small amount of text, such as word images or text blocks which contain several characters.
Therefore, most methods extract deep local features on character images and their sub-regions~\cite{nguyen2018text} or image patches~\cite{chen2019semi}.
These local features are aggregated together for computing the global feature of each handwritten page for writer identification~\cite{nguyen2018text,chen2019semi}.

Although CNNs can provide a good performance, 
there are two drawbacks to apply CNNs for writer identification based on word images: (1) the decision made by the CNN is hard to be interpreted and (2) the detailed handwriting style information on word images is not fully explored.
In order to solve these problems, we propose the FragNet method for writer identification based on word or text-block images.

\subsection{FragNet can be interpreted.} 
The proposed FragNet is inspired by Fraglets~\cite{bulacu2007text} and BagNet~\cite{brendel2019approximating}. 
Fraglets~\cite{bulacu2007text} is the grapheme-based method, which extracts the basic graphemes (sub or supra-allographic fragments) from handwritten ink traces which are characteristics of each writer.
The big advantage of Fraglets is that it can be visualized to end users.
BagNet~\cite{brendel2019approximating} provides an alternative way to understand deep learned features which are extracted on patches in input images and they are averaged for the image-level prediction, inspired by the ensemble learning~\cite{breiman1996bagging}.
The advantage of BagNet is that the importance of each part to the final classification can be visualized.
Our FragNet method can be considered as an extension of Fraglets and BagNet, which builds the neural network on each fragment in word images.
Therefore, the FragNet inherits all the advantages of Fraglets and BagNet.

Different from Fraglets~\cite{bulacu2007text}, which segments the fragment on input images, our proposed FragNet method extracts fragments (a patch with a square size) on different levels of the featurized image pyramids~\cite{lin2017feature} computed by deep convolutional neural networks.
The writer evidence of each fragment is computed by another neural network and the average of responses from fragments on the same word image is considered as the word-level writer evidence used for writer identification.

\subsection{FragNet can explore the detailed writer style information.}
\begin{figure}[!t]
    \centering
    \includegraphics[width=0.25\textwidth]{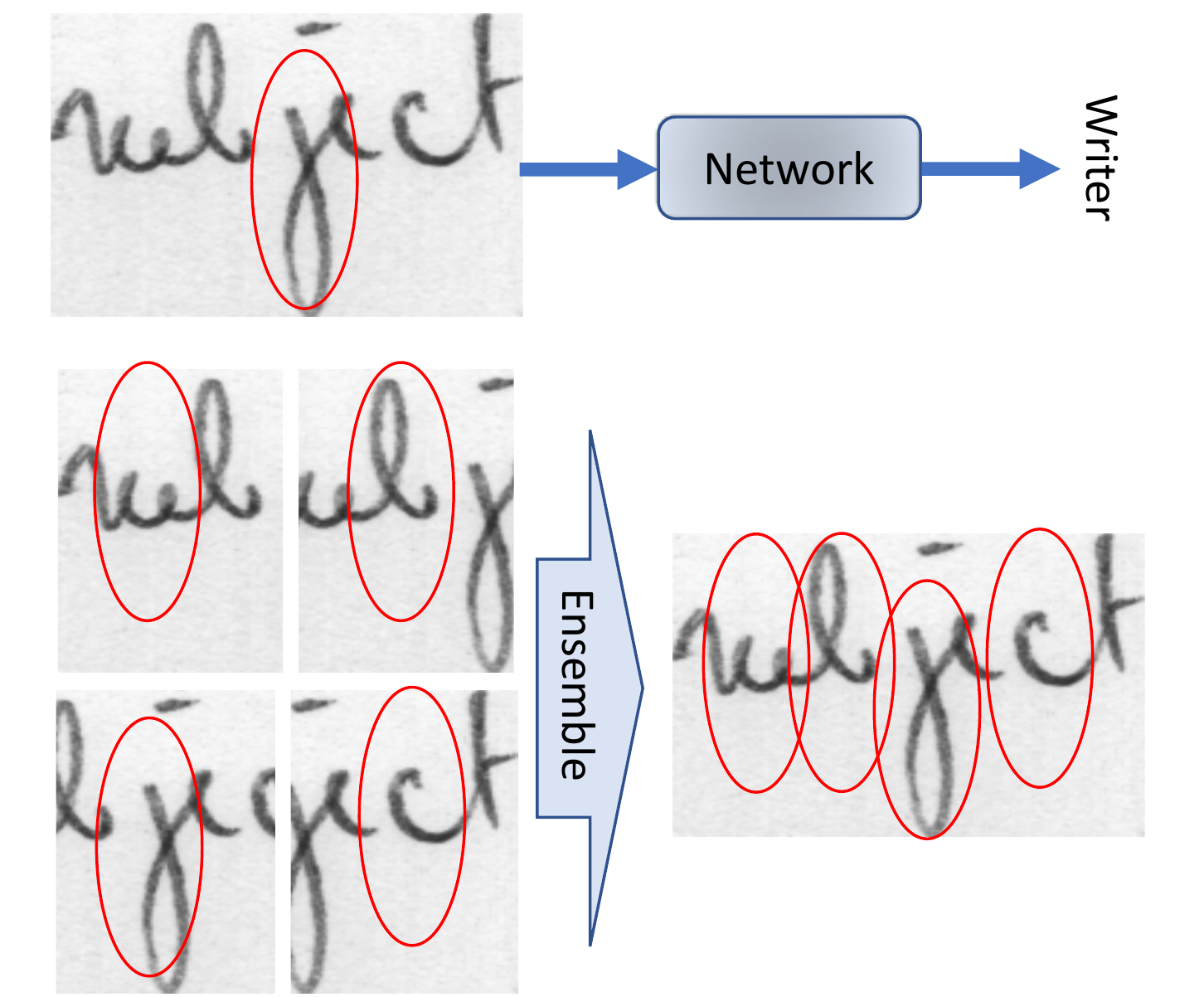}
    \caption{The red circle shows the ``spotting region" (effective receptive field) of the neural network.  FragNet has multiple spotting regions and the number of spotting regions is determined by the number of fragments. Therefore, FragNet can capture the detailed handwriting style information and provide a good performance for writer identification based on word images.}
    \label{fig:toy}
\end{figure}

Our experimental results show that the deep features learned by FragNet are much more effective for writer identification and retrieval than the deep features learned from the whole word image.
The reason is that the trained deep networks mainly focus on the most discriminative parts from the input images and ignore others~\cite{yao2019deep}. 
The discriminative part is named as ``spotting region" (effective receptive field of the network) in this paper which is similar to the attention area~\cite{jetley2018learn}. 
When training on word images, the network makes a decision based on the information from the whole image and the information in the spotting region is easy to be recognized first, resulting in a small loss during training. 
For example, when training the network to recognize the writer in Fig.~\ref{fig:toy}, the network is mainly dominated by one salient shape (the red circle means the spotting region) and other writing shapes are ignored since this salient shape yields a small loss to the network. 
However, other shapes also contain the writing style information which is useful for writer identification, thus combining their evidences properly can improve the performance. 
An efficient way to explore all shape information in the input word image is to classify and ensemble local patches, inspired by BagNet~\cite{breiman1996bagging}.
As shown in Fig.~\ref{fig:toy}, aggregating local patches results in multiple spotting regions in the input image, making the class evidence more robust.
Based on this observation, our proposed FragNet receives fragments segmented in the input image and it forces the network to learn the handwriting style information  or find the ``spotting region" in each fragment.
Combining the results of these fragments
makes the network to have multiple spotting regions, avoiding overfitting on a specific part in the input image and yielding a high performance.

The rest of this paper is organized as follows.
Section~\ref{sec:relatedwork} provides a brief summary of previous works about writer identification.
We describe the proposed FragNet in Section~\ref{sec:method} and conduct experiments in Section~\ref{sec:exps}.
The conclusion is given in Section~\ref{sec:cons}.

\section{Related Work}
\label{sec:relatedwork}

This section provides a brief review of several typical writer identification methods.
A comprehensive survey of writer identification can be found in~\cite{darganwriter}.

Before the deep learning era, handcrafted features are widely used for writer identification, which can be roughly divided into two groups: textural-based and grapheme-based features.
Based on the assumption that the writer's handwriting can be considered as a texture~\cite{said2000personal}, textural features are widely used. 
The Gabor filters are used in~\cite{said2000personal} to extract writer-specific textures.
In~\cite{helli2010text}, the Gabor and XGabor filters are used to extract features on handwritten patterns for Persian writer identification.
Newell and Griffin~\cite{newell2014writer} propose to use a bank of six Derivative-of-Gaussian filters to extract texture features at two scales.
The RootSIFT descriptors~\cite{arandjelovic2012three} are used to extract local features on handwritten images and Gaussian Mixture Models (GMM) are used for local feature encoding in~\cite{christlein2017writer,khan2018dissimilarity}. 
After obtaining the global feature representation, Christlein et al.~\cite{christlein2017writer} propose to use Exemplar-SVMs for writer identification while Khan et al.~\cite{khan2018dissimilarity} introduce the concept of similarity and dissimilarity GMM with different descriptors to compute the writing style similarity between different document images.

Model-based methods have also been widely studied for writer identification.
In~\cite{abdi2015model}, the beta-elliptic model is used to generate the grapheme codebook without training and the performance is evaluated on the Arabic handwritten documents.
Contour-based graphemes are also used in~\cite{ghiasi2013offline,schomaker2004automatic}.
Our previous work~\cite{he2015junction} uses the junction as the basic grapheme, which contains the handwriting style information.
Christlein et al.~\cite{christlein2015offline} extract local descriptors on a small image patch and use a Gaussian mixture model to encode the extracted local features into a common space for handwriting similarity measurement.
Later, they propose an unsupervised feature learning method~\cite{christlein2017unsupervised}, which learns deep features with the pseudo-label generated by $k$-means.

Several writer identification methods combine the texture and allograph features to improve the performance.
Schomaker and Bulacu~\cite{schomaker2004automatic} propose a writer identification method on the uppercase Western script using the combined features of the edge-direction and connect-component contours.
This method has been extended to the Hinge and Fraglets features in~\cite{bulacu2007text} and it has been shown in~\cite{bulacu2007text} that the combined Hinge and Fraglets features provide better performance than the individual features.
Siddiqi and Vincent~\cite{siddiqi2010text} propose two feature extraction methods: codebook-based and contour-based features. 
The small fragments of the handwritten text are extracted and mapped into a common space to generate the global feature vector.
The contour-based features describe the orientation and curvature information of the handwritten text. 
Instead of combining different types of features, the joint distribution of different attributes on writing contours also provides a good writer identification performance.
Brink et al.~\cite{brink2012writer} propose the Quill feature which is the joint distribution of the ink direction and the ink width. 
The COLD feature which is the joint distribution of the relation between orientation and length of line segments computed from ink contours is also applied for writer identification~\cite{he2017writer}.

Writer identification based on a small amount of data set has also been studied.
Adak and Chaudhuri~\cite{adak2015writer} study the writer identification based on the isolated characters and numerals.
Aubin et al.~\cite{aubin2018off} propose a writer identification method based on the simple graphemes or single strokes. 
These methods extract several handcrafted descriptors on characters or strokes and 
SVM is used for recognizing the writer identity.

Convolutional neural networks have also been studied for writer identification.
For example, methods~\cite{fiel2015writer,tang2016text} extract features from the last fully-connected layers in a trained neural network.
A multi-stream CNN is proposed in~\cite{xing2016deepwriter}, which can leverage the spatial relationship between handwritten image patches.
Nguyen et.al~\cite{nguyen2018text} train a neural network to capture local features in the whole character images and their sub-regions.
The local features extracted from tuples of images are aggregated to form the global feature of page images.
In~\cite{ni2017writer}, a denoising network is used to extract deep features on small patches.
A transfer deep learning from ImageNet is used in~\cite{rehman2019automatic} where deep features are extracted on small image patches and then fed to a SVM classifier.
Keglevic et al.~\cite{keglevic2018learning} apply a triplet network to learn a similarity measure for image patches and the global feature is computed as the vector of locally aggregated image patch descriptors.
Our previous work in~\cite{he2019deep} uses a deep adaptive learning method which learns deep features using a two-stream neural network for different tasks for writer identification based on single-word images.

\section{Approach}
\label{sec:method}

Given a set of word or text block images with writer identity, the naive way is to train a neural network on these images for writer prediction.
However, as mentioned before, the neural network usually has one spotting region in the input image, which dominates the training and the information in other regions are ignored.
In order to explore all writing style information contained in the whole word image, 
we force the network to learn the useful information on the fragments segmented on word images and feature maps computed from CNNs and combine the evidences of all fragments to make the final decision.

\subsection{Fragment segmentation in Neural Networks}
\label{sec:frags}
Fragments segmented from handwritten texts usually contain the individual's writing style information and are widely used for writer identification.
The basic assumption, as mentioned in~\cite{bulacu2007text}, is that ``the writer acts as a stochastic generator of ink-blob shapes, or graphemes".
The distribution of these segmented fragments can be used for writing style description, based on the classic bag-of-words model.
The traditional fragment-based methods need word or text segmentation.
For example, Bulacu and Schomaker~\cite{bulacu2007text} segment the word image at the minima in the lower contour of ink-trace. 
Siddiqi and Vincent~\cite{siddiqi2010text} use a square window to segment fragments on handwritten strokes.

\begin{figure}[!t]
    \centering
    \includegraphics[width=0.25\textwidth]{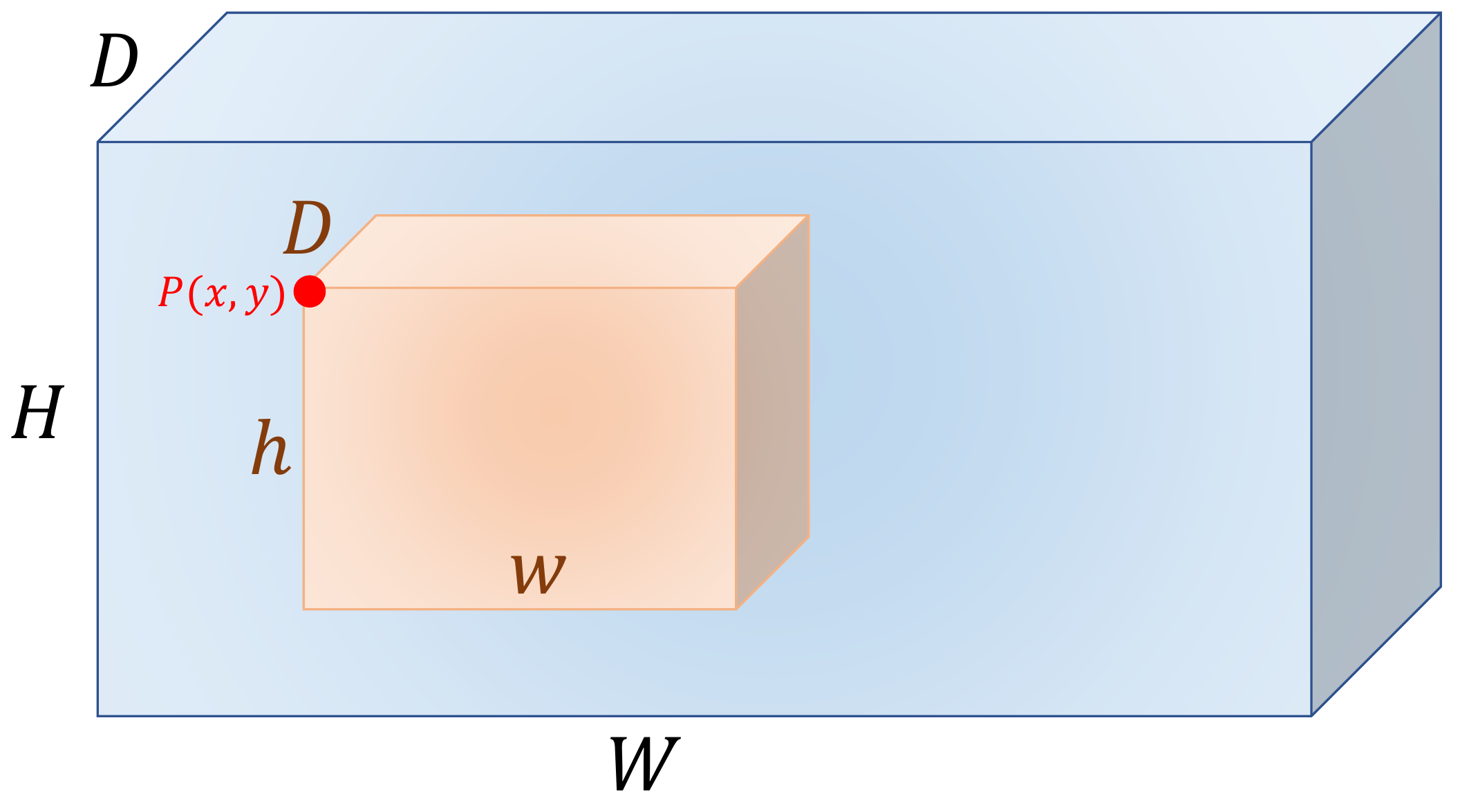}
    \caption{Fragment segmentation in a feature map space with the size: $H\times W\times D$. The fragment with the size of $h\times w\times D$ is cropped on the position $p(x,y)$.}
    \label{fig:segmentation}
\end{figure}

In this paper, we propose a different fragment segmentation method: segment the fragment in feature maps of the neural network, which contain semantic and high-level information of the handwriting style.
Our method is inspired by the featurized image pyramids~\cite{lin2017feature}, which enable a network to segment fragments on different scales by scanning on different pyramid levels.
The segmentation processing is the cutting operation, which cuts out a continuous region from a feature map of a convolutional layer, similar to the region of interest (RoI) pooling~\cite{girshick2015fast}. 
Fig.~\ref{fig:segmentation} shows an example, in which the fragment with the size of $h\times w\times D$ is segmented on the position $p(x,y)$ given the feature map with the size of $H\times W\times D$ ($h\ll H$ and $w \ll W$).
Mathematically, each fragment $A$ is defined by:
\begin{equation}
    A = \text{Crop}\big(T,\theta=(x,y,h,w)\big)
\end{equation}
where $T$ is the input tensor and the four-tuple $\theta=(x,y,h,w)$ is the set of parameters about the position and size of the fragment cropped on the input tensor $T$.
Noted that we only segment on the spatial space and
different fragments can be obtained given the different position $p(x,y)$.
The input image can be considered as a special feature map of the neural network.
Therefore, fragments can also be obtained from input images.

\subsection{FragNet networks}

\begin{figure*}[!t]
    \centering
    \includegraphics[width=\textwidth]{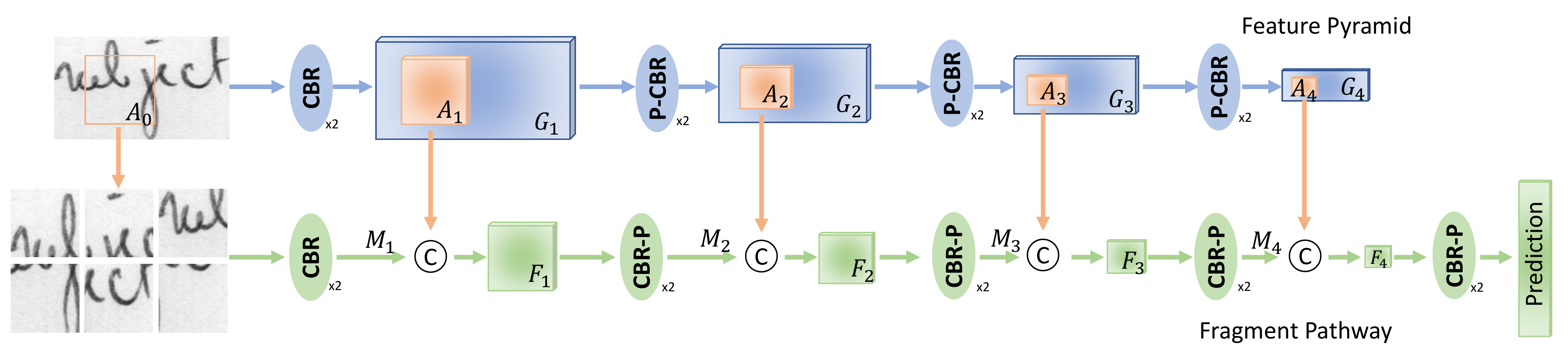}
    \caption{A FragNet network has two pathways: feature pyramid (blue color) which accepts the whole word image as the input and Fragment pathway (green color) which accepts the fragment as the input.
    (P)-CBR means the sequences of P: max-pooling, C: convolutional, B: batch normalization and R: ReLU layers. C with circle is the concatenate operation. x2 means that two blocks stacked together.
    $G_i$ and $F_i$ are the $i$-th feature maps in the feature pyramid and fragment pathway, respectively.}
    \label{fig:network}
\end{figure*}

The generic architecture of FragNet has a feature pyramid and a fragment pathway, which are fused by lateral connections. 
Fig.~\ref{fig:network} illustrates the concept of the FragNet network.

\subsubsection{Feature pyramid.} The feature pyramid (the blue pathway in Fig.~\ref{fig:network}) is the traditional convolutional neural network~\cite{krizhevsky2012imagenet} which computes the feature maps hierarchy in different scales.
It contains four $(P)$-$CBR$ blocks, in which $P$ is the max-pooling layer, $C$ is the convolutional layer, $B$ is the batch normalization layer and $R$ is the Rectified Linear Units (ReLU) layer.
The kernel size of each convolutional layer is $3\times 3$ and the stride step is 1.
The max-pooling with a kernel size of $2\times2$ and stride step of 2 is applied after every two convolutional layers in order to reduce the spatial size of the feature maps and obtain translation invariance.
The input of the feature pyramid is the whole input image, such as the word image in this paper.
Four feature maps, $G_i, i\in\{1,2,3,4\}$ as shown in Fig.~\ref{fig:network}, are obtained in the feature pyramid after every two convolutional layers:
\begin{equation}
\label{eq:conv}
    G^n_i=\circledast (G^m_{i-1},W^{m\times 3\times 3\times n}_i,W^b_i)
\end{equation}
where $\circledast$ denotes of the $P$-$CBR$ blocks, $G^1_0$ is the input image with the channel size of 1 and $W^{m\times 3\times 3\times n}_i$ is the kernel with the size of $m\times 3\times 3\times n$  in convolutional layers ($m$ is the channel size of the input feature map and $n$ is the channel size of the output feature map and $W^b_i$ is the set of parameters in the batch normalization layer.
The channel numbers of four feature maps are set to $[64,128,256,512]$, respectively.

\subsubsection{Fragment pathway.}
Fragments are segmented from the input image and feature maps on the feature pyramid.
We denote a fragment as $A_i$ which is segmented on the $i$-th feature map $G_i$ in the feature pyramid by:
\begin{equation}
    A_i = \text{Crop}\big(G_i,\theta=(x,y,h,w)\big)
\end{equation}
Specially, $A_0$ is the fragment segmented from the input image $G_0$.
The feature map of the fragment is denoted as $F_i$, which is defined as:
\begin{equation}
\begin{split}
      M^n_i&=\circledast (F^{m}_{i-1},W^{m\times 3\times 3\times n}_i,W^b_i) \\
      F_i &= [M_i,A_i] \\
\end{split}
\end{equation}
where $M_i$ is computed by $\circledast$, representing two convolutional layers $C$, followed by the Batch Normalization layer $B$ and the ReLU layer $R$.
The feature map of the fragment is denoted as $F_i$, which is the concatenation $[]$ of $M_i$ and $A_i$.
Note that $F_0=A_0$.
More details are shown in Fig.~\ref{fig:network}.

Specifically, the fragment pathway fuses the information from two directions: one is from the previous convolutional layer in the fragment pathway and one is from the fragment segmented from the corresponding convolutional layer in the feature pyramid.
Similar to DenseNet~\cite{huang2017densely}, the concatenate operation is used for merging these two different information.
The fragments are segmented in different feature maps ( $G_1,G_2,G_3,G_4$ in Fig.~\ref{fig:network}), including the input image $G_0$, by the method described in Section~\ref{sec:frags}.
In order to keep the segmented fragment consistency in the spatial space, the fragment is cropped in the same space with respect to the input image.
For example, assuming that the segment position is $p_i(x,y)$ with the size of $(h,w)$ in the feature map $G_i$ of the feature pyramid, the segment position in the next feature scale $G_{i+1}$ is $p_{i+1}(x/2,y/2)$ with the size of $(h/2,w/2)$ when the stride of the max-pooling is 2.

Finally, a global average pooling is used on the fragment pathway's output to compute the global feature vector, followed by a fully-connected classifier layer for writer prediction.
Given the input image, fragments can be segmented in different positions, yielding different fragments and predictions. 
As suggested by BagNet~\cite{brendel2019approximating}, the predictions from all fragments in one input image are averaged to infer the word-level writer evidence, which is inspired by the bagging ensemble learning~\cite{breiman1996bagging}.
Note that the parameters of the fragment pathway are shared between different fragments from the same input image.
Because there is no prior knowledge about the locations of the informative fragments, we use a sliding window strategy to segment all possible fragments on the input image and feature maps of the feature pyramid.
The stride of the sliding window on $G_0$ (the input image) is set to 16 in the vertical and horizontal directions and reduced to half after each max-pooling layer on feature maps of the feature pyramid. 

One important parameter for the fragment segmentation is the size of fragment corresponding to the input image.
In this paper, we use a square window with the size of $q\times q$ to cut the fragment in the input image.
The resulting architecture is denoted as FragNet-$q$ with the fragment size of $q\times q$ in the input image.
For each fragment, the loss is defined as the cross entropy loss between the prediction and the ground-truth, which is defined as:
\begin{equation}
    L_i = -  \sum_i^M g_i \cdot \text{log}(p_i)
\end{equation}
where $M$ is the number of writers in the training set, $g_i$ is the ground-truth of the fragment and $p_i$ is the output of the neural network after softmax.
Finally, the total training loss: $L=\sum_i^N L_{i}$ where $L_i$ is the loss of the $i$-th fragment and $N$ is the number of fragments in one word.
During testing, the writer evidence $P$ of the input word image is computed as the average response of all segmented fragments: $P=\frac{1}{N}\sum p_i$.

Our previous work in~\cite{he2019deep} for writer identification is also based on single word images. It has two branches and the information between these two branches interact at the same stages in the network.
The proposed FragNet method is different from our previous work~\cite{he2019deep} in several aspects:
(1) The aim of our previous work~\cite{he2019deep} is to use the adaptive learning to transfer features learned from different tasks while the goal of FragNet is to explore the detailed writer style information from fragments of the input word image and deep feature maps; 
(2) In~\cite{he2019deep} the two branches learn different features from different tasks while the two branches of FragNet learn features from only one task with different effective receptive fields;
(3) The input of the two branches in~\cite{he2019deep} is the same word image while the input of the feature pyramid of FragNet is the whole word image while the input of the fragment pathway is fragments segmented from the input image and deep feature maps, forcing the network of the fragment path to learn the detailed handwriting style within the fragment.

\section{Experiments}
\label{sec:exps}
In this section, we provide the experimental results of the proposed FragNet on four benchmark datasets for writer identification based on word and page images.
Unlike our previous work~\cite{he2019deep} which randomly splits word images into training and testing sets, we carefully split
all word or text block images into training and testing subsets to make sure that the word images from one page appear only in the training or the testing set, which making it possible for writer identification based on page-level images following the traditional writer identification protocol.

The word or text blocks used in this experiments are available on the author's website~\footnote{\href{https://www.ai.rug.nl/~sheng/writeridataset.html}{\url{https://www.ai.rug.nl/~sheng/writeridataset.html }}}, which can be considered as new benchmark datasets for writer identification based on word or text block images.

\subsection{Databases}

The proposed method is evaluated on four datasets: IAM~\cite{marti2002iam}, 
CVL~\cite{kleber2013cvl}, Firemaker~\cite{schomaker2000forensic} and 
CERUG-EN~\cite{he2015junction}.

\begin{table}[!t]
	\renewcommand{\arraystretch}{1.3}
	\caption{The number of training and testing word images on each data set.}
	\label{tab:dataset}
	\centering
	\begin{tabular}{l|c|c|c}
		\toprule[1pt]
		Data set & \#Writers  & \#Training & \#Testing\\
		\midrule
		IAM~\cite{marti2002iam} 				& 657 	& 56,432 	& 25,827 \\
		CVL~\cite{kleber2013cvl}				& 310 	& 62,406	& 34,564\\
		Firemaker~\cite{schomaker2000forensic}	& 250 	& 25,256	& 11,595	\\
		CERUG-EN~\cite{he2015junction}			& 105 	& 5,702		& 5,127 	\\
		\bottomrule[1pt]
	\end{tabular}
\end{table}

IAM~\cite{marti2002iam} is the widely used dataset for writer identification.
There are 1,452 pages, which are written by 657 different writers and
each writer contributes a variable number of handwritten pages.
For writers who contribute more than one page, we randomly select one page for testing and the rest pages are used for training.
For writers which contribute only one page, we randomly split text lines into training and testing sets.
The bounding boxes of word images are provided in this dataset. 
Therefore, we collect the word images from the training pages to form the training set and word images from the testing pages are used for testing. 

CVL~\cite{kleber2013cvl} contains 310 writers and each writer contributes at least 5 pages (27 writers wrote 7 pages). 
In this paper, the first three pages are used for training and the rest pages are used for testing.
Similar to the IAM dataset, word images are also available on this dataset. 

The Firemaker dataset~\cite{schomaker2000forensic} contains handwritten documents from 250 Dutch subjects and each writer contributes four pages.
Similar to~\cite{bulacu2007text}, we use the text images in Page 1 for training and the text images in Page 4 for testing.
The CERUG-EN dataset~\cite{he2015junction} contains handwritten documents from 105 Chinese subjects with two paragraphs in English.
In this paper, we use the first paragraph for training and the second paragraph for testing.
Since there is no word box existed in the Firemaker and CERUG-EN datasets, we roughly segment these documents into text blocks (most text blocks contain only one word).

\begin{figure}
	\centering
	\includegraphics[width=0.4\textwidth]{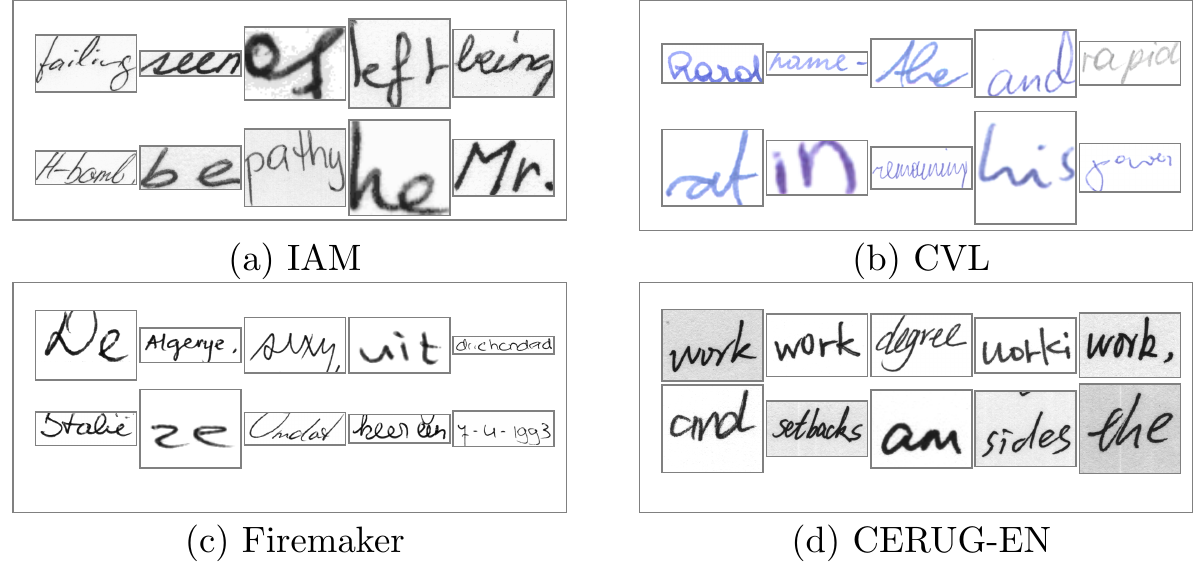}
	\caption{Several examples of training samples on each data set.}
	\label{fig:examples}
\end{figure}

All word images are resized to a fixed size $(64,128)$ by keeping the aspect ratio without distortions with white-pixel padding until reaching the pre-determined size.
No augmentation method is used during training.
Table~\ref{tab:dataset} shows the number of training and testing samples and Fig.~\ref{fig:examples} shows several training samples on each dataset.
From the figure we can see that our training samples are based on word regions, which could contain at least two characters.

\subsection{Implementation details}
Neural networks are trained by the Adam optimizer~\cite{kingma2014adam}.
The mini-batch size is set to 10 due to the limitation of the GPU memory.
The initial learning rate is set to 0.0001 and reduce to half at the epoch 10 and 20 and the network is fine-tuned with a learning rate of 0.00001 at the last 5 epochs.
The whole training takes 30 epochs.
In this paper, we evaluate FragNet-$q$ with $q\in[64,32,16]$.

All experiments are conducted on one GPU card (Tesla V100-SXM2) with 16GB RAM using Tensorflow.
We use the FLOPs (floating point operations) to measure the computational complexity. 
The FLOPs  of a convolutional operation is $c_i\times h\times w\times k_i\times k_j\times c_o$ where $c_i$ and $c_o$ are the dimensions of input and output tensors, $h$ and $w$ are the height and width of the tensor, $k_i$ and $k_j$ are the kernel size.
There are eight convolutional layers on the feature pyramid and the dimensions of feature maps are [64, 128, 256, 512] in different scales.
Thus, the FLOPs of WordImgNet is around 1.05G. The FLOPs of FragNet-64, FragNet-32 and FragNet-16 are around 7.14G, 7.41G and 3.90G, respectively.

\subsection{Performance of writer identification based on word images}
In this section, we evaluate the proposed method for writer identification based on word or text block images. 
We compare FragNet with two baselines: WordImgNet and ResNet18~\cite{he2016deep}. WordImgNet is the network with the same structure as the fragment pathway of FragNet, receiving the whole word image as the input and producing one global prediction.
For ResNet18~\cite{he2016deep}, we use the kernel size of 3$\times$3 at the first convolutional layer and remove the max-pooling layer to keep the detailed writing style information on the earlier stage.
Our proposed FragNet, however, receives the whole word images and fragments segmented from the input image and feature maps of the feature pyramid.
For each word image, several fragments are segmented and fed into the trained FragNet to identify the writer. 
In the evaluation, we take the average response of different fragments segmented from the input word image as the final prediction of the same input word image.
The training configurations of WordImgNet, ResNet18 and FragNet are the same for fair comparison.

\subsubsection{Comparison with baselines}
Table~\ref{tab:allographsize} shows the writer identification performance of two baselines: WordImgNet, ResNet18~\cite{he2016deep} and the proposed FragNet method.
From the table we can see that FragNet-$q$ with a larger fragment size ($q$) provides better results on the four datasets, which is similar to the conclusion found in BagNet~\cite{brendel2019approximating}.
In addition, the proposed FragNet-$64$ provides better results than the baselines WordImgNet and ResNet18 in terms of the Top-1 performance. The FragNet-$32$ gives better results than the baseline WordImgNet in the IAM and CVL data sets while worse results in the Firemaker and CERUG-EN data sets.
The results suggest the effectiveness of the proposed FragNet-$64$ for writer identification based on word images.

\begin{table}[!t]
    \centering
    \caption{Performance of writer identification with different networks based on \textbf{word images}.}
    \label{tab:allographsize}
    \resizebox{0.5\textwidth}{!}{
    \begin{tabular}{c|cc|cc|cc|cc}
    \toprule[1pt]
         \multirow{2}{*}{Net} & \multicolumn{2}{c|}{IAM} & \multicolumn{2}{c|}{CVL} & \multicolumn{2}{c|}{Firemaker} & \multicolumn{2}{c}{CERUG-EN}\\
         \cline{2-9}
         &  Top-1 & Top-5 &  Top-1 & Top-5 &  Top-1 & Top-5 &  Top-1 & Top-5 \\
         \hline
         ResNet18~\cite{he2016deep} & 83.2 & 94.3 & 88.5 & 96.7 & 63.9 & 86.4 & 70.6 & 94.0  \\
         WordImgNet & 81.8 & 94.1 & 88.6 & 96.8 & 67.9 & 88.1 & 77.3 & \textbf{96.4}\\
         FragNet-$16$ & 79.8 & 93.3 & 89.0 & 97.2 & 59.6 & 83.2 & 60.6 & 90.3\\
         FragNet-$32$ & 83.6 & 94.8 & 89.0 & 97.3 & 65.0 & 86.8 & 62.3 & 90.1\\
         FragNet-$64$ & \textbf{85.1} & \textbf{95.0} & \textbf{90.2} & \textbf{97.5} & \textbf{69.0} & \textbf{88.5} & \textbf{77.5} & 95.6\\
         \bottomrule[1pt]
    \end{tabular}}
\end{table}

\subsubsection{Performance of writer identification with different word lengths}
\begin{figure}[!t]
    \centering
    \includegraphics[width=0.22\textwidth]{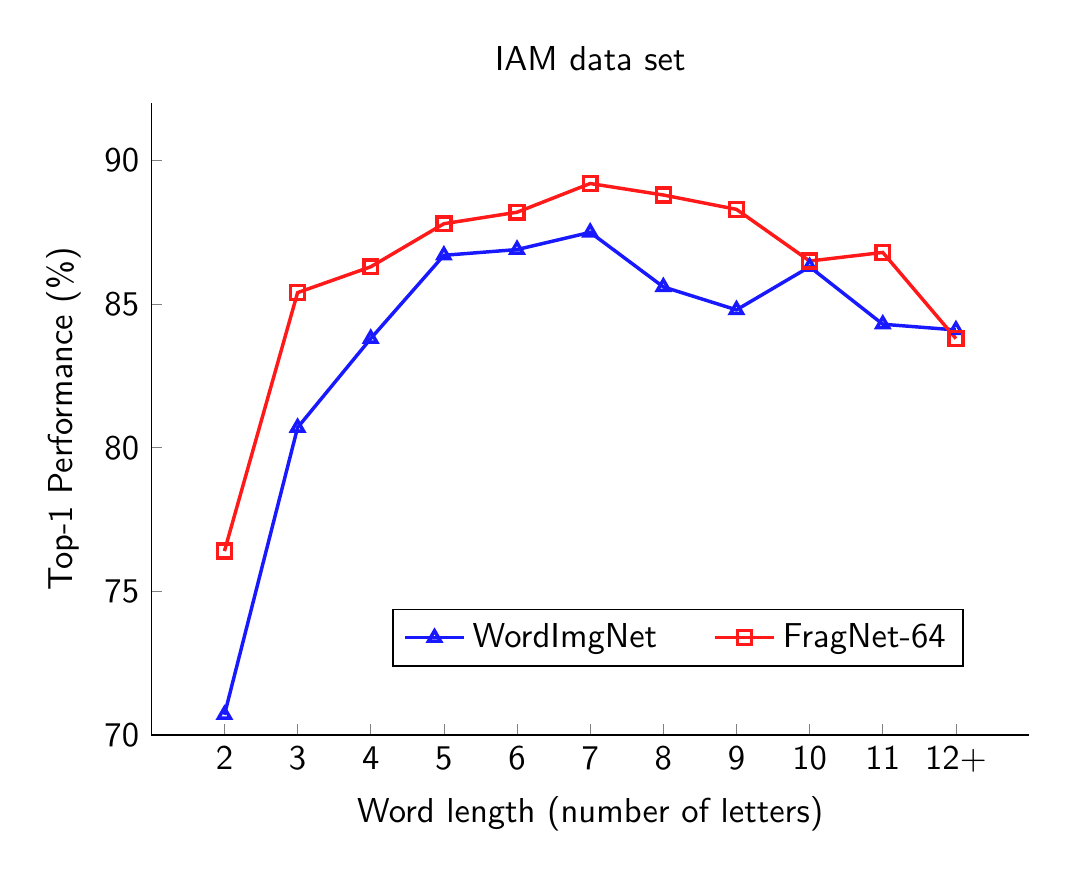}
    \includegraphics[width=0.22\textwidth]{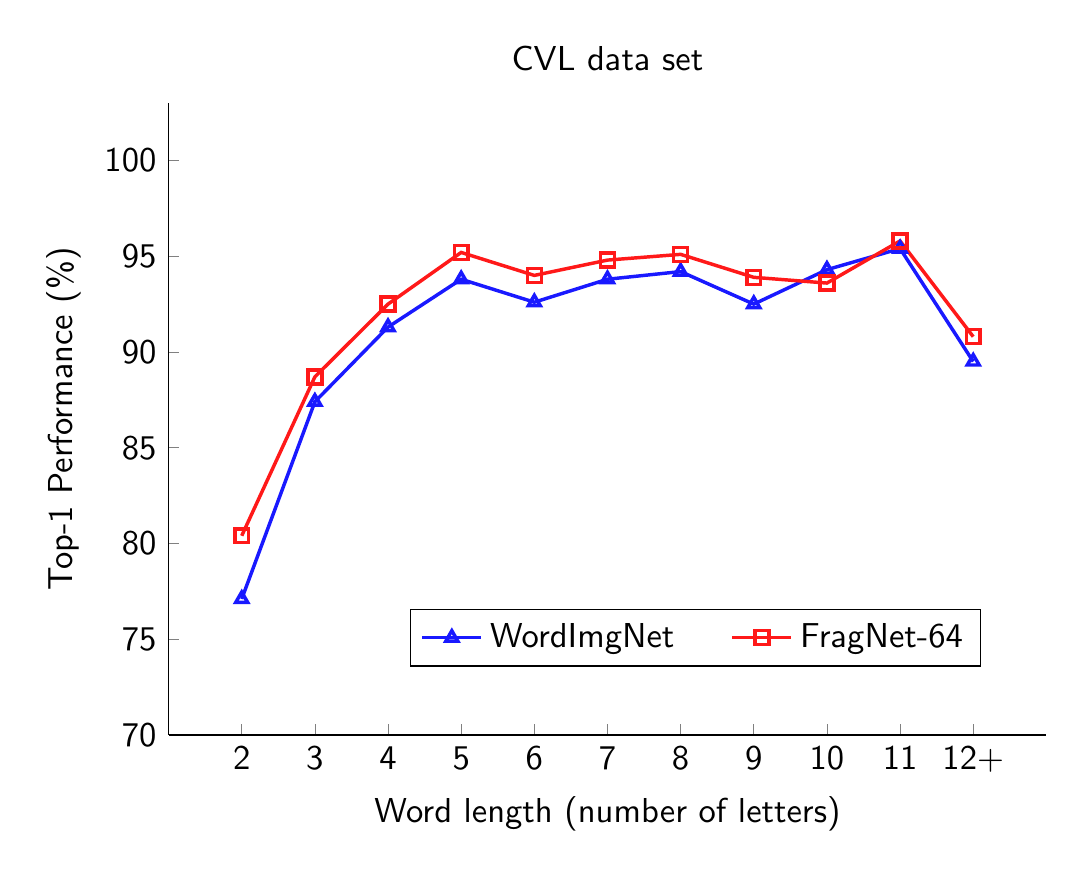}
    \caption{Performance of writer identification (Top-1) for FragNet-$64$ and WordImgNet with different word lengths on the IAM and CVL data sets.}
    \label{fig:wordlen}
\end{figure}

Fig.~\ref{fig:wordlen} shows the Top-1 performance of writer identification with different word lengths on the IAM and CVL datasets since each word image in these two data sets carries the word label. 
Similar to our previous work~\cite{he2019deep}, word images with only two characters contain less texts and thus the performance of both FragNet-$64$ and WordImgNet is lower than the performance of other words. 
The performance of word images with more then three letters is usually high because there are more allograhpic information in these words which are sufficient for training.
Moreover, FragNet-$64$ provides better results than WordImgNet with different word lengths, especially on the word images with only two letters. 

\subsubsection{Accuracy distribution between FragNet-$64$ and WordImgNet}
\begin{figure*}
    \centering
    \includegraphics[width=\textwidth]{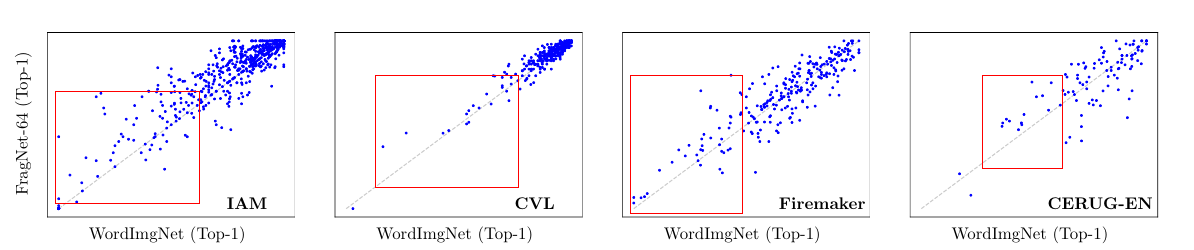}
    \caption{Scatter plots of the writer-conditional Top-1 accuracy of FragNet-$64$ against WordImgNet in four datasets. Note that the number of dots in these figures equals to the number of writers in the corresponding dataset. The range of both the x- and y-axis are [0,100]. The red boxes show the most writers with low performance given by WordImgNet and their performance is improved by the proposed FragNet-$64$. }
    \label{fig:errdis}
\end{figure*}

In Fig.~\ref{fig:errdis}, we plot the Top-1 performance within each writer in the four data sets of FragNet-$64$ against the performance of WordImgNet.
We can see that the accuracy distribution is fairly consistent between FragNet-$64$ and WordImgNet. 
The performance of FragNet-$64$ is low on the writers whose performance of WordImgNet is also low.
However, FragNet-$64$ can improve the performance and provide a high performance (the red boxes in Fig.~\ref{fig:errdis}) on these writers whose performance given by WordImgNet is low, especially on the IAM dataset.

\subsubsection{Visualization of the informative fragments}
\begin{figure}[!t]
    \centering
    \includegraphics[width=0.5\textwidth]{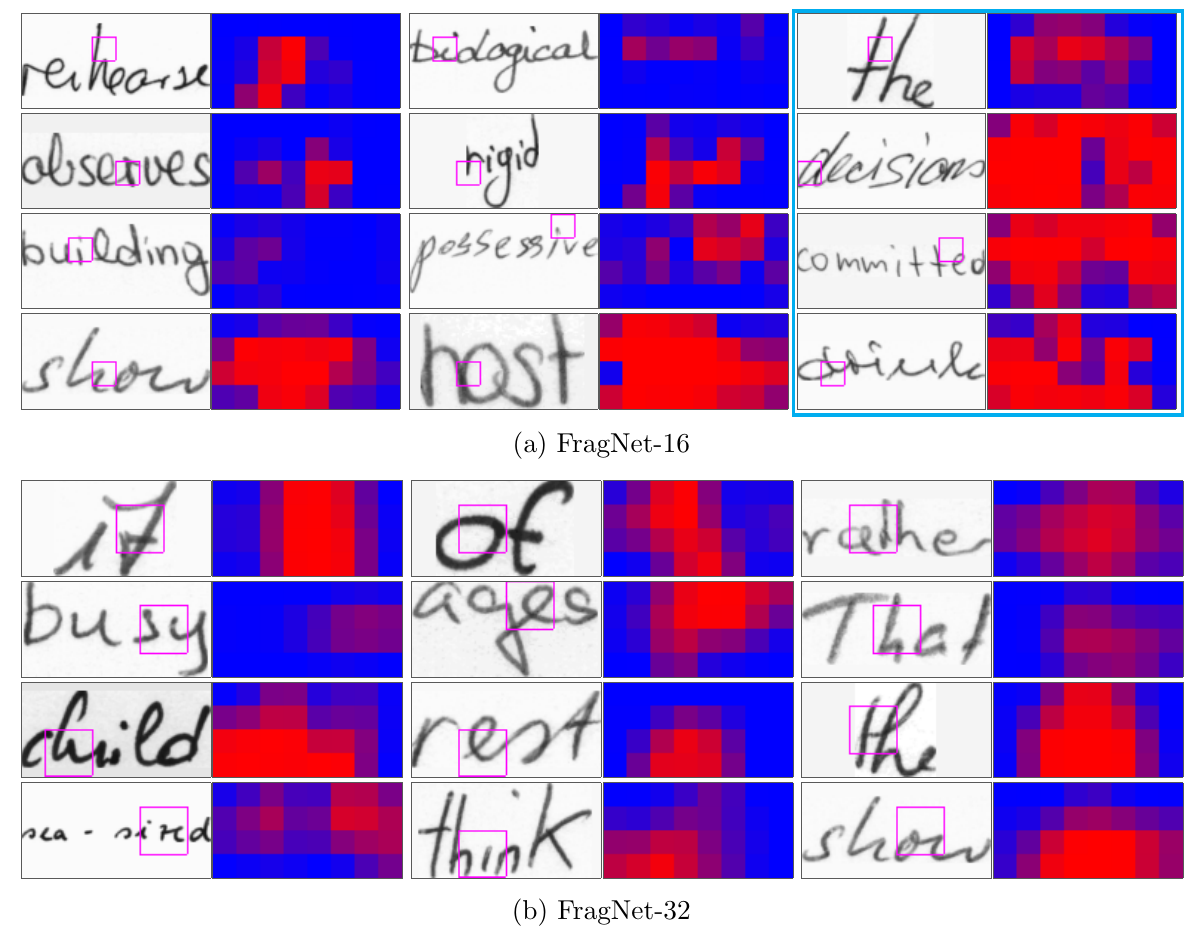}
    \caption{The word images and the corresponding heatmaps of FragNet-$16$ and FragNet-$32$ on the IAM dataset. The rectangle in each word image is the most predictive fragment. The heatmap of each word shows the writer evidence extracted from each fragment. The spatial sum over the evidence is the total writer evidence. 
     Failure cases (last column in Fig.(a) with the cyan box): Note that the prediction will not only rely on fragments segmented in the input image, but also rely on fragments segmented in feature maps of the feature pyramid. Therefore, some fragments which contain a small part of texts or no text can still provide a high evidence for writer identification since they also receive the global information on the feature pyramid.}
    \label{fig:fragNets32}
\end{figure}

Since the writer evidence is from the input image and fragments, which making it possible to visualize which fragment combined with the word image contributes most to the writer identity.
In this section, we display these heatmaps of FragNet-$16$ and FragNet-$32$ for the predicted writer evidence on word images from the IAM dataset in Fig.~\ref{fig:fragNets32}.
Highlighting the most informative fragments is very useful for explanation.

\subsubsection{Comparison with the handcrafted features}
We evaluate the effectiveness of features learned by the WordImgNet and FragNet-$64$ networks for writer identification, comparing with the traditional handcrafted features.
For FragNet, the features of fragments on the input word image are averaged to form the final feature representation.
Given features extracted on word images, the writer model is built by the averaging of word features in the training set from the same writer. 
The writer identification is performed by the nearest neighbor method, which is widely used in writer identification~\cite{bulacu2007text,siddiqi2010text}.
Table~\ref{tab:compareOthers} shows the results of writer identification based on word images.
From the table we can see that the performance of the handcrafted features for writer identification based on word images is very low since these textural features capture the statistics information of handwritten text, which usually requires a large amount of text to obtain a stable representation.
However, the features learned by neural networks provide much better results than the handcrafted features and the proposed FragNet-$64$ gives the best performance.

\begin{table}[!t]
	\centering 
	\renewcommand{\arraystretch}{1.3}
	\caption{The comparison of writer identification performance based on \textbf{word images} using different features.}
	\label{tab:compareOthers}
	\resizebox{0.5\textwidth}{!}{
	\begin{tabular}{c|cc|cc|cc|cc}
		\toprule[1pt]
		\multirow{2}{*}{Method} & \multicolumn{2}{c|}{IAM} & \multicolumn{2}{c|}{CVL} & \multicolumn{2}{c|}{Firemaker} & \multicolumn{2}{c}{CERUG-EN} \\
		\cline{2-9}
													& Top1 & Top5 & Top1 & Top5 & Top1 & Top5 & Top1 & Top5 \\
		 \midrule
		Hinge~\cite{bulacu2007text} 				& 13.8 & 28.3 & 13.6 & 29.7 & 19.6 & 40.0 & 14.4 & 32.8\\
		Quill~\cite{brink2012writer}				& 23.8 & 44.0 & 23.8 & 46.7 & 21.7 & 43.7 & 24.5 & 51.9 \\
		CoHinge~\cite{he2017beyond}					& 19.4 & 34.1 & 18.2 & 34.2 & 27.4 & 48.2 & 17.7 & 34.0\\
		QuadHinge~\cite{he2017beyond} 				& 20.9 & 37.4 & 17.8 & 35.5 & 26.5 & 47.4 & 17.0 & 36.0\\
		COLD~\cite{he2017writer}					& 12.3 & 28.3 & 12.4 & 29.0 & 22.7 & 45.1 & 17.3 & 42.2 \\
		Chain Code Pairs~\cite{siddiqi2010text} 	& 12.4 & 27.1 & 13.5 & 30.3 & 17.5 & 36.8 & 14.5 & 33.0\\
		Chain Code Triplets~\cite{siddiqi2010text}	& 16.9 & 33.0 & 17.2 & 35.4 & 22.9 & 43.8 & 17.8 & 38.0\\
		\midrule 
		WordImgNet &  52.4 & 70.9 & 62.5 & 82.0 & 50.2 & 75.8 & 74.3 & 94.6\\
		FragNet-$64$ & \textbf{72.2} & \textbf{88.0} & \textbf{79.2} & \textbf{93.3} & \textbf{57.5} & \textbf{80.8} & \textbf{75.9} & \textbf{94.7}\\
		\bottomrule
	\end{tabular}}
\end{table}

\subsubsection{Performance of writer retrieval}
We evaluate the effectiveness of the deep learned features by WordImgNet and FragNet-$64$ for writer retrieval on the test set.
Writer retrieval is similar to the writer identification problem, which aims to find the word images which are written by the same writer of the query.
If there is only one ground truth existed in the data set given the query, writer retrieval is exactly the same as the writer identification problem using the ``leave-one-out" strategy~\cite{bulacu2007text,siddiqi2010text}.
Each word image on the test set is tested against all remaining ones on the test set and the performance of writer retrieval is evaluated by the mean average precision (mAP) and Top-1 rate, similar to~\cite{christlein2017writer,zheng2015scalable}.

Table~\ref{tab:retrieval} shows the results of writer retrieval using deep features extracted from WordImgNet and FragNet-$64$.
The proposed FragNet-$64$ provides better results than WordImgNet, which demonstrates that the proposed FragNet-$64$ can learn a robust and effective deep feature representation for writer retrieval.

\begin{table}[!t]
    \centering
    \caption{Performance of writer retrieval with features extracted on different networks based on \textbf{word images}.}
    \label{tab:retrieval}
    \resizebox{0.5\textwidth}{!}{
    \begin{tabular}{c|cc|cc|cc|cc}
    \toprule[1pt]
         \multirow{2}{*}{Net} & \multicolumn{2}{c|}{IAM} & \multicolumn{2}{c|}{CVL} & \multicolumn{2}{c|}{Firemaker} & \multicolumn{2}{c}{CERUG-EN}\\
         \cline{2-9}
         &  Top-1 & mAP &  Top-1 & mAP &  Top-1 & mAP &  Top-1 & mAP \\
         \hline
         WordImgNet & 79.4 & 0.3011 & 80.9 & 0.2813 & 60.7 & 0.2386& 80.0 & 0.5017\\
         FragNet-$64$ & 86.5 & 0.4947 & 87.1 &0.4939  & 65.4  & 0.3310 & 80.0 & 0.5544 \\
         \bottomrule[1pt]
    \end{tabular}}
\end{table}

\begin{table}[!t]
    \centering
    \caption{Performance of writer identification with different networks based on \textbf{page images}.}
    \label{tab:pageres}
    \resizebox{0.5\textwidth}{!}{
    \begin{tabular}{c|cc|cc|cc|cc}
    \toprule[1pt]
         \multirow{2}{*}{Net} & \multicolumn{2}{c|}{IAM} & \multicolumn{2}{c|}{CVL} & \multicolumn{2}{c|}{Firemaker} & \multicolumn{2}{c}{CERUG-EN}\\
         \cline{2-9}
         &  Top-1 & Top-5 &  Top-1 & Top-5 &  Top-1 & Top-5 &  Top-1 & Top-5 \\
         \hline
         %ResNet~\cite{he2016deep} & 81.7 & 93.9 & 87.4 & 96.6 & 61.5  & 84.3 & 62.4 & 91.9\\
         WordImgNet & 95.8 & \textbf{98.0} & 98.8 & \textbf{99.4} & \textbf{97.6} & 98.8 & 97.1 &\textbf{100}\\
         FragNet-$16$ & 94.2 & 97.4 & 98.5 & \textbf{99.4} & 92.8 & 98.0 & 79.0 & 97.1\\
         FragNet-$32$ & 95.3 & \textbf{98.0} & 98.6 & \textbf{99.4} & 96.0 & 99.2 & 84.7 & 97.1 \\
         FragNet-$64$ & \textbf{96.3} & \textbf{98.0} & \textbf{99.1} & \textbf{99.4} & \textbf{97.6} & \textbf{99.6} & \textbf{98.1} & \textbf{100}\\
         \bottomrule[1pt]
    \end{tabular}}
\end{table}

\begin{figure}[!t]
    \centering
    \includegraphics[width=0.22\textwidth]{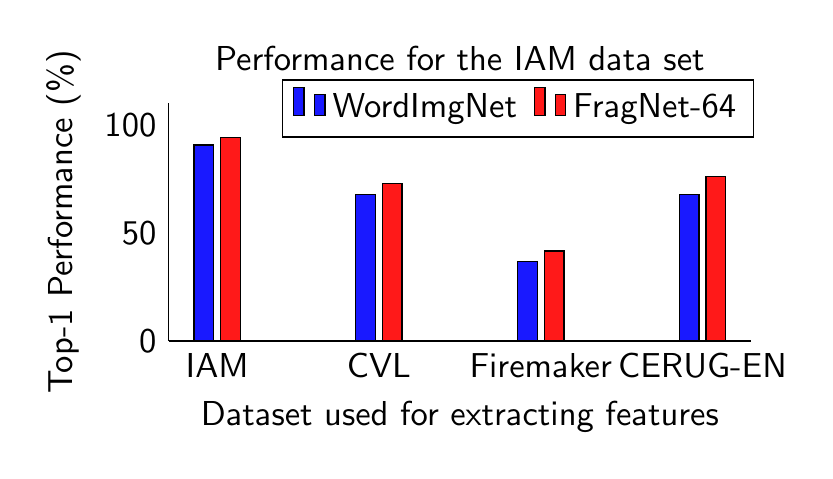}
    \includegraphics[width=0.22\textwidth]{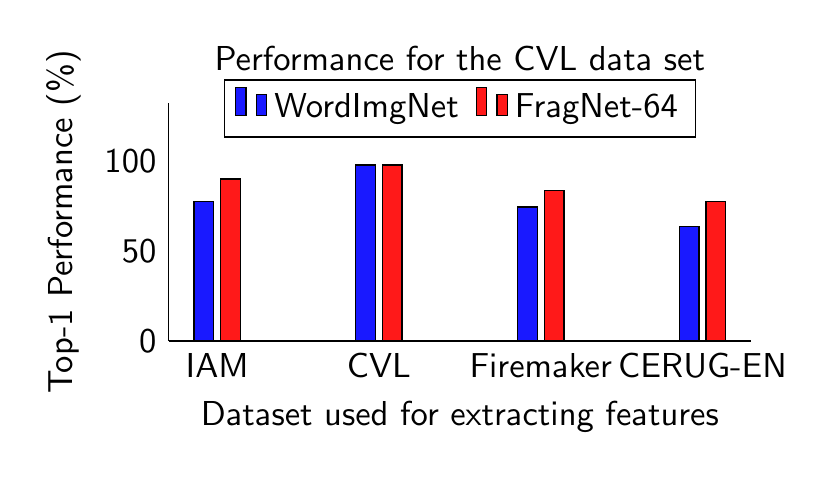}
    \includegraphics[width=0.22\textwidth]{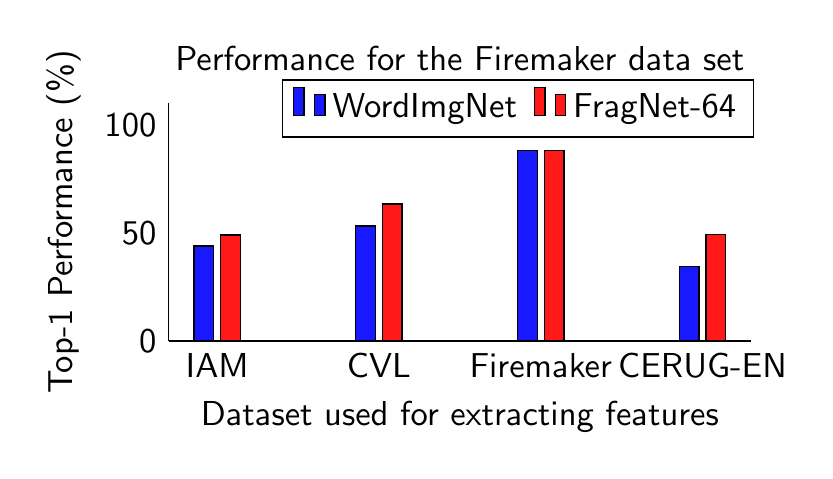}
    \includegraphics[width=0.22\textwidth]{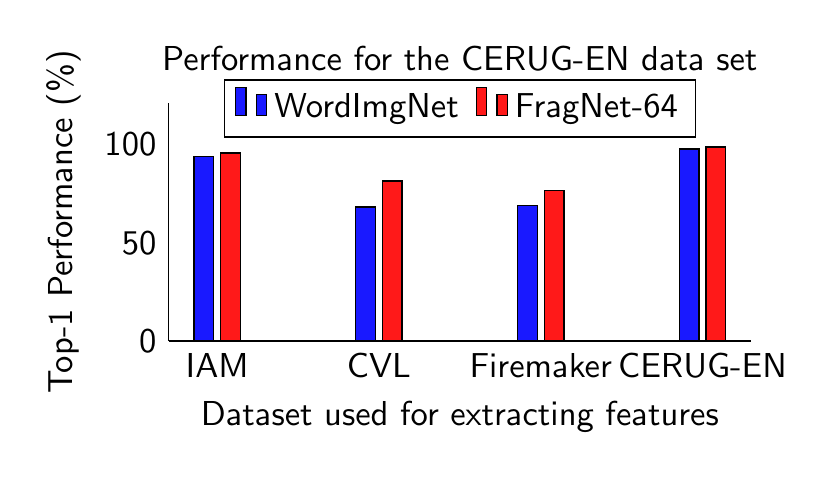}
    \caption{The performance of writer identification based on page images by features extracted from neural networks trained on different data sets.}
    \label{fig:crossset}
\end{figure}

\subsection{Performance of writer identification based on page images}
Following the traditional methods which perform writer identification based on page images, we also evaluate the proposed method for writer identification based on the global feature extracted on the whole page.
We evaluate the proposed method in two scenarios:
within-dataset and cross-datasets evaluation.

\subsubsection{Within-dataset evaluation}
In this section, we conduct the within-dataset evaluation for writer identification. The writer evidence of each test page is computed by averagely aggregating word features extracted by the network trained with training samples from the same dataset:
\begin{equation}
    P_{page} = \frac{1}{N} \sum_{w\in page}^N P(w)
\end{equation}
where $P_{page}$ is the writer probability on the $page$ and $P(w)$ is the writer probability of the word $w$ from the network. $N$ is the total number of words on the $page$.
Table~\ref{tab:pageres} shows the performance of deep features extracted from WordImgNet and FragNet for writer identification based on the page level. 
Compared to Table~\ref{tab:allographsize}, the performance of writer identification on page images is much better than the performance on word images.
In addition, the proposed FragNet-$64$ provides slight better results than others in these four datasets in terms of the Top-1 performance.

 \subsubsection{Cross-datasets evaluation} 
Given a trained network, the global feature vector of one page is the average aggregation of deep features of all word images within the page.
 Once feature vectors of all pages on one dataset are extracted, writer identification is performed by the widely used ``leave-one-out" strategy.
There are many different methods for extracting the global features by aggregating word image features in one page. 
However, as found in~\cite{nguyen2018text}, the average aggregation provides the best performance. 
Therefore, in this paper, we apply the average aggregation of all word images in one page to compute the global feature for writer identification.
The dimension of the feature vector is equal to the number of writers on each data set since we extract the deep feature on the last classification layer.

 We provide the performance of deep learned features extracted from networks trained using different datasets for writer identification based on page images.
 The global feature of each page on one dataset is computed as the average of word features extracted from the network trained on other datasets.
Fig.~\ref{fig:crossset} shows the experimental results of WordImgNet and FragNet-$64$ on different datasets.
For example, the top-left panel shows the performance of writer identification on the IAM dataset, using deep features trained on the training set of the IAM, CVL, Firemaker and CERUG-EN datasets. 
From the figure we can see that the best performance on each dataset is given by deep features extracted on the neural network trained on training samples from the same dataset. The reason is that the handwriting styles of different datasets are quite different and thus the trained neural network can only capture the handwriting style which appears in the training dataset.
In addition, the proposed FragNet-$64$ provides better results than WordImgNet, which indicates that the FragNet can capture the detailed handwriting style information which does not appear in the training set.

Moreover, the handwriting style on the IAM and CERUG-EN datasets are similar since features extracted from the network trained on the CERUG-EN dataset also provide a good performance on the IAM dataset and vice versa. 
However, the handwriting style on the Firemaker dataset whose handwritten documents are wrote in Dutch is quite different from others.

\begin{table}[!t]
    \centering
    \caption{Comparison of state-of-the-art methods on the IAM dataset.}
    \label{tab:IAMcomps}
    \resizebox{0.5\textwidth}{!}{
    \begin{tabular}{lclc}
    \toprule[1pt]
         Reference & \#Writer & Feature & Top-1(\%)  \\
         \hline
        Siddiqi and Vincent~\cite{siddiqi2010text} & 650 & Contour and codebook features & 91.0 \\
        He and Schomaker~\cite{he2017beyond} & 650 & Best results among 17 handcraft features & 93.2 \\
        Khalifa et al.~\cite{khalifa2015off} & 650 & Graphemes with codebook & 92.0 \\
        Hadjadji and Chibani~\cite{hadjadji2018two} & 657 & LPQ, RL and oBIF with OC-K-Means & 94.5 \\
        Wu et al.~\cite{wu2014offline} & 657 & Scale invariant feature transform & \textbf{98.5} \\
        Khan et al.~\cite{khan2018dissimilarity} &  650 & SIFT+RootSIFT & 97.8 \\
        Nguyen et. al~\cite{nguyen2018text} & 650 & CNN with sub-images of size $64\times 64$ & 93.1 \\
        WordImgNet & 657 & CNN with word images & 95.8 \\
        FragNet-$64$ & 657 & CNN with word images and fragments & 96.3 \\
         \bottomrule[1pt]
    \end{tabular}}
\end{table}

\begin{table}[!t]
    \centering
    \caption{Comparison of state-of-the-art methods on the CVL dataset.}
    \label{tab:CVLcomps}
    \resizebox{0.5\textwidth}{!}{
    \begin{tabular}{lclc}
    \toprule[1pt]
         Reference & \#Writer & Feature & Top-1(\%)  \\
         \hline
         Fiel and Sablatnig~\cite{fiel2013writer} & 309 & SIFT with GMM & 97.8 \\
        Tang and Wu~\cite{tang2016text} & 310 & CNN with joint Bayesian & \textbf{99.7} \\
        Christlein et al.~\cite{christlein2017writer} & 310 & SIFT with GMM and Examplar-SVMs & 99.2 \\
        Khan et al.~\cite{khan2018dissimilarity} &  310 & SIFT+RootSIFT & 99.0 \\
        WordImgNet & 310 & CNN with word images & 98.8 \\
        FragNet-$64$ & 310 & CNN with word images and fragments & 99.1 \\
         \bottomrule[1pt]
    \end{tabular}}
\end{table}

\begin{table}[!t]
    \centering
    \caption{Comparison of state-of-the-art methods on the Firemaker dataset.}
    \label{tab:Firemakercmps}
    \resizebox{0.5\textwidth}{!}{
    \begin{tabular}{lclc}
    \toprule[1pt]
         Reference & \#Writer & Feature & Top-1(\%)  \\
         \hline
         He and Schomaker~\cite{he2017beyond} & 250 & Best results among 17 handcraft features & 92.2 \\
         Wu et al.~\cite{wu2014offline} & 250 & Scale invariant feature transform & 92.4 \\
        Nguyen et. al~\cite{nguyen2018text} & 250 & CNN with sub-images of size $64\times 64$ & 93.6 \\
        Khan et al.~\cite{khan2018dissimilarity} &  250 & SIFT+RootSIFT & \textbf{98.0} \\
        WordImgNet & 250 & CNN with word images & 97.6 \\
        FragNet-$64$ & 250 & CNN with word images and fragments & 97.6 \\
         \bottomrule[1pt]
    \end{tabular}}
\end{table}

\begin{table}[!t]
    \centering
    \caption{Comparison of state-of-the-art methods on the CERUG-EN dataset.}
    \label{tab:cerugcomps}
    \resizebox{0.5\textwidth}{!}{
    \begin{tabular}{lclc}
    \toprule[1pt]
         Reference & \#Writer & Feature & Top-1(\%)  \\
         \hline
        He and Schomaker~\cite{he2017beyond} & 105 & Best results among 17 handcraft features & 97.1 \\
        WordImgNet & 105 & CNN with word images & 98.1 \\
        FragNet-$64$ & 105 & CNN with word images and fragments & \textbf{100.0} \\
         \bottomrule[1pt]
    \end{tabular}}
\end{table}

\subsubsection{Comparison with other studies}
In this section, we summarize state-of-the-art methods about writer identification in the literature in Tables~\ref{tab:IAMcomps},~\ref{tab:CVLcomps},~\ref{tab:Firemakercmps},~\ref{tab:cerugcomps}.
Please note  that these methods are not comparable to our method because they follow different experimental protocols. Nevertheless, we can draw some conclusions.
Generally, deep learned features provide better results than the best handcrafted features shown in~\cite{he2017beyond}.
Our proposed method is comparable to other state-of-the-art methods.
Especially, our method is better than the system~\cite{nguyen2018text}, which also extracts deep features on handwritten patches with a size of $64\times 64$. 
The difference is that the patches in~\cite{nguyen2018text} are randomly selected in handwritten documents while fragments in the proposed method are uniformly segmented from word images and feature maps on the feature pyramid.

\section{Conclusion}
\label{sec:cons}

This paper proposed a FragNet network for writer identification based on fragments segmented on the input image and feature pyramid maps in the convolutional neural network.
Writer identification is evaluated based on word or text block images and page images. 
We shown that networks trained with word images and fragments (the proposed FragNet) can provide better performance than networks trained with only word images (the baseline WordImgNet) for the word-based and page-based writer identification on four benchmark datasets.
Generally, deep features learned from fragments with a large size provide better performance.
Specifically, FragNet-$64$ provides much better results than WordImgNet on word images which contains only two letters (Fig.~\ref{fig:wordlen}) and it can improve the performance on these writers whose performance given by WordImgNet is low (Fig.~\ref{fig:errdis}).
More importantly, the decision made by FragNet is based on input images and fragments, thus the most informative fragments can be visualized. 
Note that the aim of the proposed method is not to achieve state-of-the-art performance, but to show that the fragment based network can improve the performance and can be visualized for end users.
In addition, the proposed FragNet could be used for on-line writer identification with a sliding window strategy.

One of the limitations of the proposed FragNet is that it needs word image or region segmentation, which is challenging on the highly cursive writing documents.
This limitation leads to a direction of future works, focusing on extending FragNet for writer identification on any handwritten document (such as historical documents~\cite{fiel2017icdar2017}) without segmentation.

\bibliographystyle{IEEEtran}
\bibliography{IEEEabrv,main}

\end{document}